\documentclass[]{article}
\usepackage{times}
\usepackage[left=1in, right=1in, top=1in, bottom=1in]{geometry}
\usepackage{breakcites}

\usepackage{cite}
\usepackage{amsmath,amssymb,amsfonts}
\usepackage{graphicx}
\usepackage{textcomp}
\usepackage{xcolor}
\def\BibTeX{{\rm B\kern-.05em{\sc i\kern-.025em b}\kern-.08em
    T\kern-.1667em\lower.7ex\hbox{E}\kern-.125emX}}
    
\usepackage{bm}
\usepackage{mathtools}
\usepackage{algorithm}
\usepackage{algpseudocode}
\DeclareMathOperator{\tr}{\mathrm{Tr}} 
\DeclareMathOperator{\diag}{\mathrm{diag}} 
\usepackage{hyperref}
\usepackage{multicol}
\usepackage{tabu}
\usepackage{multirow}
\usepackage{pbox}
\usepackage{ctable} %
\usepackage{subcaption}
\usepackage{adjustbox}
\usepackage{placeins}
\usepackage{wrapfig}
\usepackage{comment}
\usepackage{pbox}

\makeatletter
\newcommand{\printfnsymbol}[1]{%
  \textsuperscript{\@fnsymbol{#1}}%
}
\makeatother

\begin{document}

\title{Robust Generative Restricted Kernel Machines using Weighted Conjugate Feature Duality}

\author{Arun Pandey\thanks{Authors contributed equally to this work.}, Joachim Schreurs\printfnsymbol{1} \& Johan A. K. Suykens\\
Department of Electrical Engineering ESAT-STADIUS, KU Leuven \\
Kasteelpark Arenberg 10, B-3001 Leuven, Belgium \\
\{arun.pandey, joachim.schreurs, johan.suykens\}@esat.kuleuven.be}

\maketitle

\begin{abstract}
Interest in generative models has grown tremendously in the past decade.
However, their training performance can be adversely affected by contamination, where outliers are encoded in the representation of the model. This results in the generation of noisy data. In this paper, we introduce weighted conjugate feature duality in the framework of Restricted Kernel Machines (RKMs). The RKM formulation allows for an easy integration of methods from classical robust statistics. This formulation  is used to fine-tune the latent space of generative RKMs using a weighting function based on the Minimum  Covariance  Determinant, which  is  a  highly  robust  estimator of  multivariate location and scatter. Experiments show that the weighted RKM is capable of generating clean images when contamination is present in the training data. We further show that the robust method also preserves uncorrelated feature learning through qualitative and quantitative experiments on standard datasets.

\end{abstract}

\section{Introduction}
A popular choice for generative models in machine learning are latent variable models such as Variational Auto-Encoders (VAE) \cite{kingma_auto-encoding_2013}, Restricted Boltzmann Machines (RBM) \cite{Smolensky:1986, salakhutdinov_deep} and Generative Adversarial Networks (GAN)~\cite{goodfellow_generative_2014,arjovsky2017wasserstein,liu2016coupled}. These latent spaces provide a representation of the input data by embedding into an underlying vector space. Exploring these spaces allows for deeper insights in the structure of the data distribution, as well as understanding relationships between data points. The interpretability of the latent space is enhanced when the model learns a disentangled representation~\cite{higgins2017beta,chen2016infogan}. In a disentangled representation, a single latent feature is sensitive to changes in a single generative factor, while being relatively invariant to changes in other factors \cite{bengio2013representation}. For example hair color, lighting conditions or orientation of faces. 

In generative modelling, training data is often assumed to be ground truth, therefore outliers can severely degrade the learned representations and performance of trained models. 
The same issue arises in generative modelling where contamination of the training data results in encoding of the outliers. Consequently, the network generates noisy images when reconstructing out-of-sample extensions. To solve this problem, multiple robust variants of generative models were proposed in~\cite{futami2017variational,chrysos2018robust,tang2012robust}. However, these generative models require clean training data or only consider the case where there is label noise. In this paper, we address the problem of \emph{contamination on the training data itself}. This is a common problem in real-life datasets, which are often contaminated by human error, measurement errors or changes in system behaviour. To the best of our knowledge, this specific problem is not addressed in other generative methods. The Restricted Kernel Machine (RKM) formulation~\cite{suykens_deep_2017} allows for a straightforward integration of methods from classical robust statistics to the RKM framework. The RKM framework yields a representation of kernel methods with visible and hidden units establishing links between kernel methods~\cite{suykens_least_2002} and RBMs. \cite{joachim} showed how kernel PCA fits into the RKM framework. A tensor-based multi-view classification model was developed in \cite{houthuys_tensor-based_nodate}. In \cite{GENRKM}, a multi-view generative model called Generative RKM (Gen-RKM) is introduced which uses explicit feature-maps for joint feature-selection and subspace learning. Gen-RKM learns the basis of the latent space, yielding uncorrelated latent variables. This allows to generate data with specific features, i.e. a disentangled representation. 

\textbf{Contributions:} This paper introduces a weighted Gen-RKM model that detects and penalizes the outliers to regularize the latent space. Thanks to the introduction of weighted conjugate feature duality, a RKM formulation for weighted kernel PCA is derived. This formulation is used within the Gen-RKM training procedure to fine-tune the latent space using different weighting schemes. A weighting function based on Minimum Covariance Determinant (MCD)~\cite{rousseeuw1999fast} is proposed. Qualitative and quantitative experiments on standard datasets show that the proposed model is unaffected by large contamination and can learn meaningful representations.

  \begin{figure*}[!t]
    \centering
    	\begin{subfigure}{0.292\textwidth}
		\centering
		\includegraphics[trim={1.28cm 0.1cm 1cm 0cm},clip, width=0.95\textwidth]{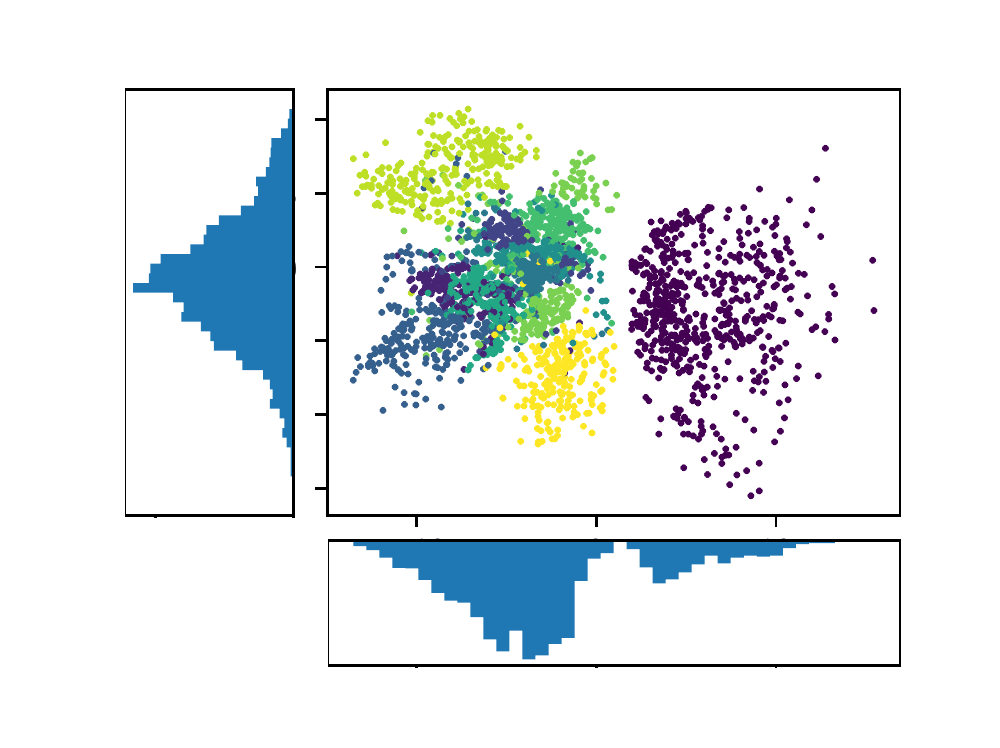} 
		\caption{VAE}
	\end{subfigure}
    \begin{subfigure}{0.292\textwidth}
		\centering
		\includegraphics[trim={0.5cm 0.1cm 1cm 0cm},clip,width=1\textwidth]{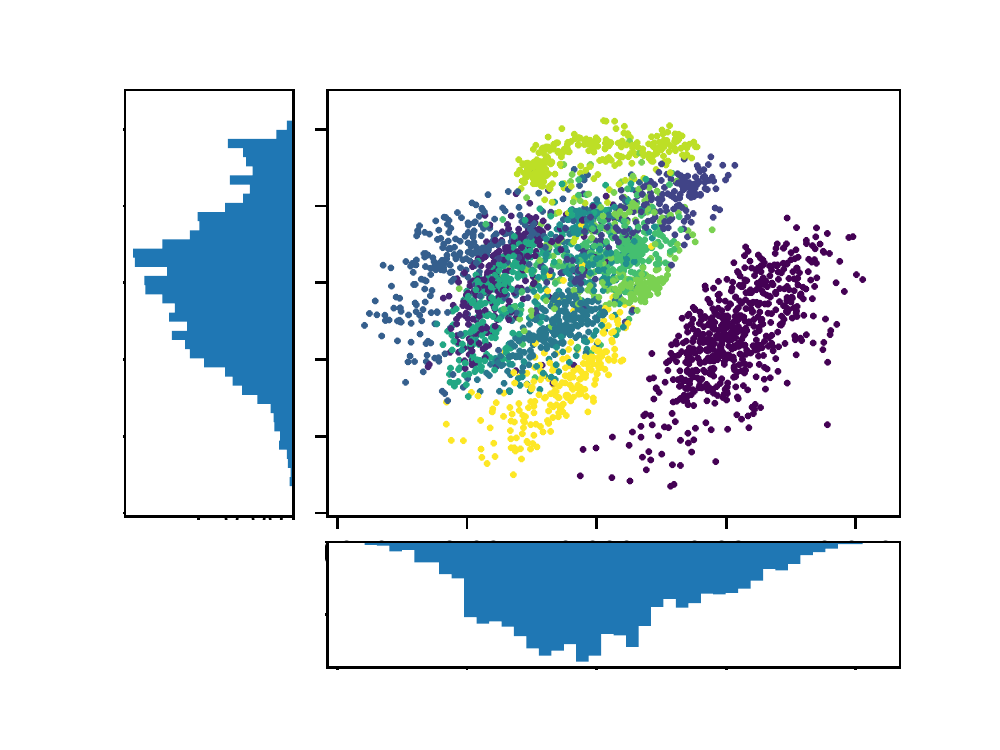}
		\caption{Gen-RKM}
	\end{subfigure}
	\begin{subfigure}{0.292\textwidth}
		\centering
		\includegraphics[trim={0.5cm 0.1cm 1cm 0cm},clip,width=1\textwidth]{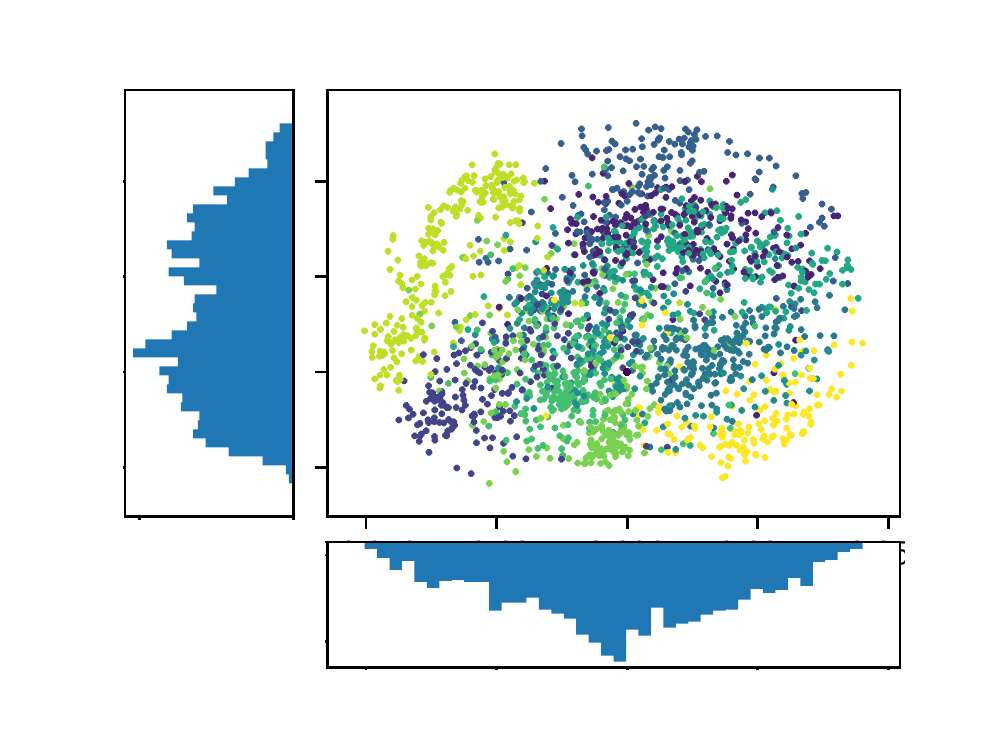} 
		\caption{Robust Gen-RKM}
	\end{subfigure}
	\begin{subfigure}{0.1\textwidth}
		\raggedright
		\includegraphics[trim={0.1cm -1cm 0.1cm 0cm},clip,height=3cm]{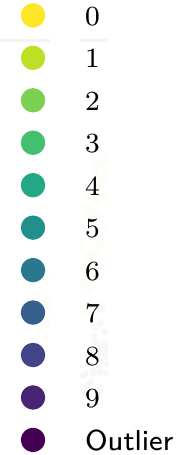}
	\end{subfigure}
	\caption[]{Illustration of robustness against outliers on the MNIST dataset. 20\% of the training data is contaminated with noise. The models are trained with a 2-dimensional latent space in the standard setup, see Section \ref{sec:Experiments}. The presence of outliers distorts the distribution of the latent variables for the Gen-RKM and VAE, where the histogram of the latent variables is skewed. By down-weighting the outliers, the histogram resembles a Gaussian distribution again.}
    \label{fig:histogramOutliers}
\end{figure*}
\section{Weighted Restricted Kernel Machines}
\label{sec:weightedRKM}
\subsection{Weighted Conjugate Feature Duality}
For a comprehensive overview of the RKM framework, the reader is encouraged to refer~\cite{suykens_deep_2017,GENRKM}.
In this section, we extend the notion of conjugate feature duality by introducing a weighting matrix.
Assuming $\bm{D} \succ 0$ to be a positive-definite diagonal weighting matrix, the following holds for any two vectors $\bm{e}, \bm{h}\in \mathbb{R}^{n}$, $\lambda>0$:
\begin{equation}\label{eq:weighted_fenchel}
    \frac{1}{2\lambda}\bm{e}^{\top}\bm{D}\bm{e} + \frac{\lambda}{2}\bm{h}^{\top}\bm{D}^{-1}\bm{h}\geq \bm{e}^{\top}\bm{h}.
\end{equation}
The inequality could be verified using the Schur complement by writing the above in its quadratic form:
\begin{equation}
    \frac{1}{2}\begin{bmatrix}
        \bm{e}^{\top} & \bm{h}^{\top} 
    \end{bmatrix}
    \begin{bmatrix}
        \frac{1}{\lambda}\bm{D}\bm{I} & -\bm{I} \\
        -\bm{I} & \lambda \bm{D}^{-1}\bm{I} 
    \end{bmatrix}
    \begin{bmatrix}
        \bm{e}\\ \bm{h} 
    \end{bmatrix}      \geq 0.  
\end{equation}
It states that for a matrix $\bm{Q} = \left[\begin{smallmatrix}  \bm{A} & \bm{B} \\ \bm{B}^{\top} & \bm{C} \end{smallmatrix}\right],$ one has $\bm{Q} \succeq 0$ if and only if $\bm{A} \succ 0$ and the Schur complement $\bm{C} - \bm{B}^{\top} \bm{A}^{-1} \bm{B} \succeq 0$ \cite{Boyd:2004:CO:993483}, which proves the above inequality. This is also known as the Fenchel-Young inequality for quadratic functions \cite{rockafeller1987}.\\

We assume a dataset $\mathcal{D}=\{ \bm{x}_{i}\}_{i=1}^{N}$, with $ \bm{x}_{i} \in \mathbb{R}^d $ consisting of $N$ data points. For a feature-map $\phi: \mathbb{R}^{d}\mapsto \mathbb{R}^{d_{f}}$ defined on input data points, the weighted kernel PCA objective~\cite{scholkopf1997kernel} in the Least-Squares Support Vector Machine (LS-SVM) setting is given by~\cite{alzate2008multiway}:
\begin{equation}
 \begin{aligned}
    \min_{\bm{U},\bm{e}} J(\bm{U},\bm{e})  =   ~ \frac{\eta}{2}\tr(\bm{U}^{\top}\bm{U}) - \frac{1}{2\lambda} \bm{e}^{\top}\bm{D}\bm{e} \enskip
     \text{s.t.} \enskip \bm{e}_i=\bm{U}^{\top}\phi(\bm{x}_i), \forall i=1,\ldots,N,
\end{aligned}
\end{equation}
where $ \bm{U} \in \mathbb{R}^{d_f \times s} $ is the unknown interconnection matrix. By using \eqref{eq:weighted_fenchel}, the error variables $\bm{e}_{i}$ are conjugated to latent variables $\bm{h}_{i}\in \mathbb{R}^{s}$ and substituting the constraints into the objective function yields
\begin{equation}\label{eq:obj_train_weighted}
    J\leq{\mathcal{J}_{t}^{\bm{D}}\coloneqq \sum_{i=1}^{N} \left\{-\phi(\bm{x}_{i})^\top \bm{U}\bm{h}_i + \frac{\lambda}{2} \bm{D}_{ii}^{-1}\bm{h}_{i}^\top \bm{h}_i\right\} + \frac{\eta}{2}\tr(\bm{U}^\top \bm{U})}.
\end{equation}
The stationary points of $\mathcal{J}_{t}^{\bm{D}}$ are given by:
\begin{equation}\label{eq:obj_train_weighted_Station}
    \begin{cases}
        \frac{\partial \mathcal{J}_{t}^{\bm{D}}}{\partial \bm{h}_{i}}=0  \implies & {\lambda \bm{D}_{ii}^{-1} \bm{h}_i =  \bm{U}^\top \phi(\bm{x}_i) }, \enskip \forall i= 1,\dots, N \\
        \frac{\partial \mathcal{J}_{t}^{\bm{D}}}{\partial \bm{U}}=0     \implies  & {\bm{U} = \frac{1}{\eta} \sum_{i=1}^{N}\phi(\bm{x}_{i})\bm{h}_{i}^\top }.
    \end{cases}
\end{equation}
Eliminating $\bm{U}$ and denoting the kernel matrix $\bm{K} \coloneqq [k(x_i,x_j)]_{ij}$ with kernel function $k(x,y) = \langle \phi(\bm{x}), \phi(\bm{y})\rangle$, the eigenvectors $\bm{H}\coloneqq [\bm{h}_{1},\dots,\bm{h}_{N}],~ \bm{\Lambda} \coloneqq \diag\{\lambda_1,\ldots,\lambda_s\}\in\mathbb{R}^{s\times s} $ such that $\lambda_1 \geq \ldots \geq \lambda_s$ with $s$ the dimension of the latent space, we get the weighted eigenvalue problem:
\begin{equation}
\label{eq:weightedEIG}
    \frac{1}{\eta}\bm{[DK] H}^{\top} = \bm{H}^{\top}\bm{\Lambda}.
\end{equation} 
One can verify that each eigenvalue-eigenvector pair lead to the value $\mathcal{J}_{t}^{\bm{D}} = 0$. Using the weighted kernel PCA potential outliers can be penalized, which is discussed  in more detail in Section~\ref{sec:weightingScheme}.
\begin{figure*}[!t]
    \centering
    \begin{subfigure}{0.3\textwidth}
		\centering
		\includegraphics[width=0.8\textwidth]{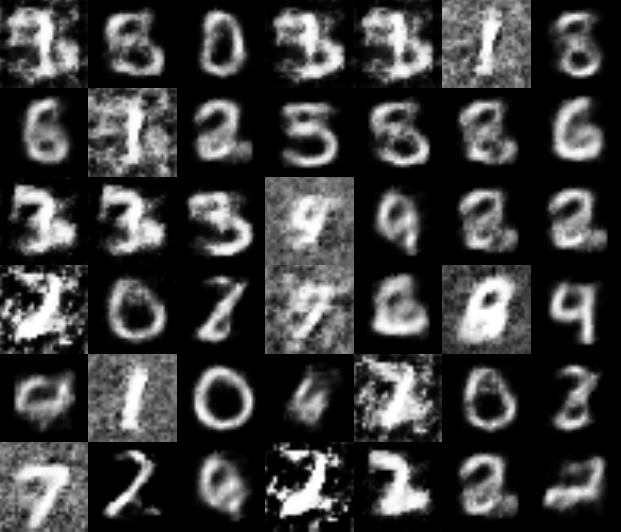}
        \caption{VAE}
	\end{subfigure}
    \begin{subfigure}{0.3\textwidth}
		\centering
		\includegraphics[width=0.8\textwidth]{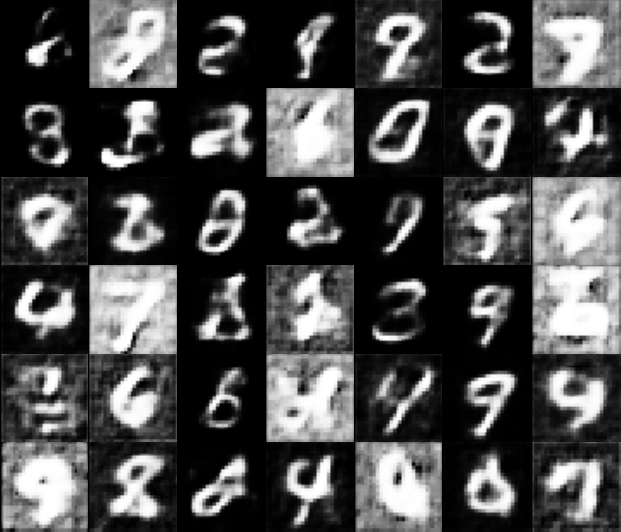}
		\caption{Gen-RKM}
	\end{subfigure}
	\begin{subfigure}{0.3\textwidth}
		\centering
		\includegraphics[width=0.8\textwidth]{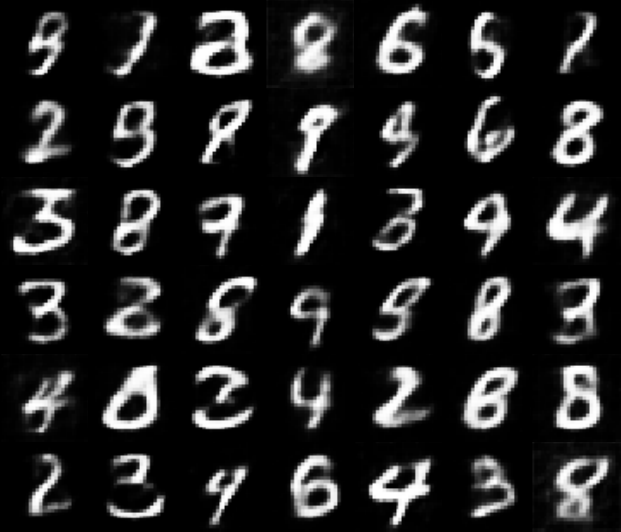}
        \caption{Robust Gen-RKM}
	\end{subfigure}
	\caption[]{Illustration of robust generation on  the MNIST dataset. 20\% of the training data is contaminated with noise. The images are generated by random sampling from a fitted Gaussian distribution on the learned latent variables. When using a robust training procedure, the model does not encode the noisy images. As a consequence, no noisy images are generated.}
    \label{fig:RandomGeneration}
\end{figure*}
\subsection{Generation} 
Given the learned interconnection matrix $\bm U$, and a latent variable $\bm{h}^{\star}$, consider the following objective function
\begin{equation}\label{eq:obj_gen_lin}
    \mathcal{J}_{g}= - \phi(\bm{x}^{\star})^{\top}\bm{U}\bm{h}^{\star}  + \frac{1}{2}\phi(\bm{x}^{\star})^{\top}\phi(\bm{x}^{\star}),
\end{equation}
with a regularization term on the input data. Here ${\mathcal{J}_{g}}$ denotes the objective function for generation. To reconstruct or denoise a training point, $\bm{h}^{\star}$ can be one-of-the corresponding hidden units of the training point. Random generation is done by fitting a normal distribution on the learned latent variables, afterwards a random $\bm{h}^{\star}$ is sampled from the distribution which is put through the decoder network. Note that in the training objective  (ref. \eqref{eq:obj_train_weighted}) we are imposing a soft-Gaussian prior over latent variables through quadratic regularization on $\{\bm{h}_i\}_{i=1}^{N}$. The stationary points of~\eqref{eq:obj_gen_lin} yields the \emph{generated feature vector}~\cite{GENRKM,joachim}  $\varphi(\bm{x}^{\star})$, given by the corresponding $\bm{h}^{\star}$. With slight abuse of notation, we denote the generated feature-vector by  $ {\varphi}(\bm{x}^{\star}) = [ \frac{1}{\eta} \sum_{i=1}^{N}{\phi}(\bm{x}_i)\bm{h}_{i}^\top ] \bm{h}^{\star}$, which is a point in the feature-space corresponding to an unknown $\bm{x}^{\star}$ in data space.
To obtain the generated data in the input space, the inverse image of the feature map $\phi(\cdot)$ should be computed. In kernel methods, this is known as the  pre-image problem. We seek to find the function $\psi \colon \mathbb{R}^{d_f}  \mapsto \mathbb{R}^{d}$, such that $(\psi \circ \varphi) (\bm{x}^\star)\approx \bm{x}^\star$, where $\varphi (\bm{x}^\star)$ is calculated from above. The pre-image problem is known to be ill-conditioned~\cite{mika_kernel_nodate}, and consequently various approximation techniques have been proposed~\cite{honeine_preimage_2011-1}. Another approach is to explicitly define pre-image maps and learn the parameters in the training procedure~\cite{GENRKM}. In the experiments, we use (convolutional) neural networks as the feature maps $\phi_{\bm{\theta}}(\cdot)$, where the notation extends to $\varphi_{\bm{\theta}}(\cdot)$. Another (transposed convolutional) neural network is used for the pre-image map $\psi_{\bm{\zeta}}(\cdot)$ \cite{dumoulin2016guide}. The parameters $\bm{\theta}$ and $\bm{\zeta}$ correspond to the network parameters. These parameters are learned by minimizing the reconstruction error in combination with the weighted RKM objective function. The training algorithm is described in more detail in section \ref{sec:algorithm}.  

\noindent \emph{Remark on Out-of-Sample extension}: To reconstruct or denoise an out-of-sample test point $\bm{x}^{\star}$, the data is projected on the latent space using:
\begin{equation}
\bm{h}^{\star} = \lambda^{-1} \bm{U}^\top \phi(\bm{x}^{\star}) =  \frac{1}{\lambda \eta} \sum_{i=1}^{N} \bm{h}_{i} k(\bm{x}_{i},\bm{x}^{\star}).
\end{equation}
The latent point is reconstructed by projecting back to the input space by first computing the generated feature vector followed by its pre-image map $\psi_{\bm \zeta}(\cdot)$.

\section{Robust estimation of the latent variables}
\label{sec:weightingScheme}
\subsection{Robust Weighting Scheme}
In this paper, we propose a weighting scheme to make the estimation of the latent variables more robust against contamination. The weighting matrix is a diagonal matrix with a weight $\bm{D}_{ii}$ corresponding to every $\bm{h}_i$ such that:
\begin{equation}
\label{eq:weights}
\bm{D}_{ii}=\left\{\begin{array}{ll}
{1} & {\text { if }} d_{i}^2 \leq \chi_{s, \alpha}^{2}\\
{10^{-4}} & {\text { otherwise, }}
\end{array}\right.
\end{equation}
with $s$ the dimension of the latent space, $\alpha$ the significance level of the Chi-squared distribution and $d_{i}^2$ the Mahalanobis distance for the corresponding $\bm{h}_i$:
\begin{equation}
 d_{i}^2 = \left(\bm{h}_i-\hat{\bm{\mu}}\right)^{\top} \hat{\bm{S}}^{-1}\left(\bm{h}_i-\hat{\bm{\mu}}\right),
\end{equation}
with $\hat{\bm{\mu}}$ and $\hat{\bm{S}}$ the robustly estimated mean and covariance matrix respectively. In this paper, we propose to use the Minimum Covariance Determinant (MCD)~\cite{rousseeuw1999fast}. The MCD is a highly robust estimator of multivariate location and scatter which has been used in many robust multivariate statistical methods~\cite{hubert2005robpca, hubert2012deterministic}. Given a data matrix of $N$ rows with $s$ columns, the objective is to find the $N_\mathrm{{MCD}} < N$ observations whose sample covariance matrix has the lowest determinant. Its influence function is bounded \cite{croux1999influence} and has the highest possible breakdown value when $N_\mathrm{{MCD}} = \lfloor (N + s + 1)/2 \rfloor$. 
In the experiments, we typically take $N_\mathrm{{MCD}} = \lfloor N \times 0.75 \rfloor$ and $\alpha = 0.975$ for the Chi-squared distribution. The user could further tune these parameters according to the estimated contamination degree in the dataset. Eventually, the reweighting procedure can be repeated iteratively, but in practice one single additional weighted step will often be sufficient. 
\begin{figure*}[!t]
    \centering
  \includegraphics[width=0.9\textwidth]{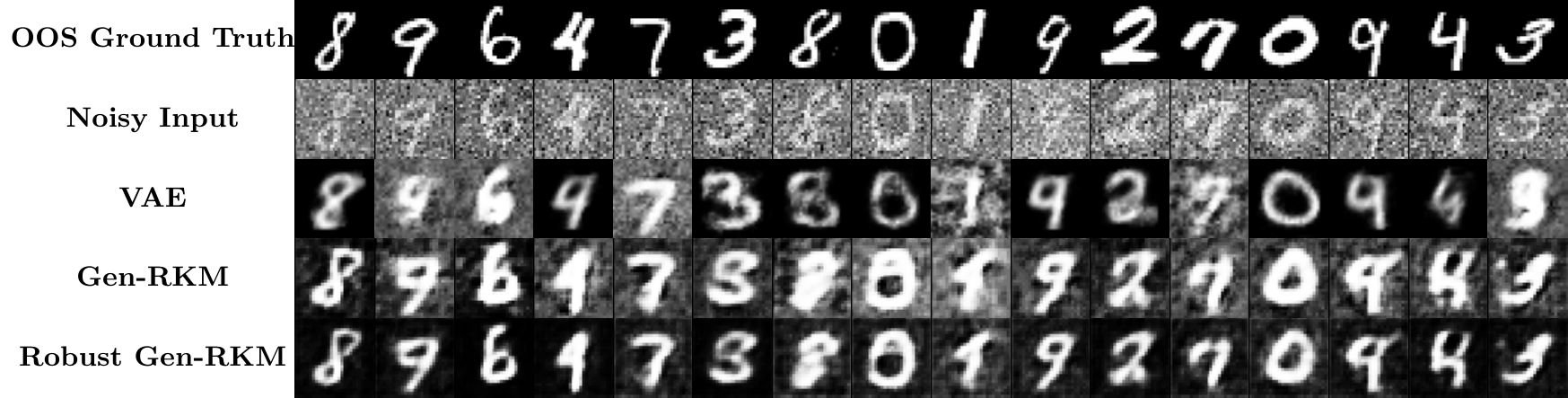}
    \caption[]{Illustration of robust denoising on the MNIST dataset. 20 \% of the training data is contaminated with noise. The first and second row show the clean and noisy test images respectively. The third, fourth and fifth row show the denoised image using the VAE, Gen-RKM and robust Gen-RKM  respectively.}
    \label{fig:Densoing}
\end{figure*}
Kernel PCA can take the interpretation of a one-class modeling problem with zero target value around which one maximizes the variance \cite{suykens2003support}. The same holds in the Gen-RKM framework. This is a natural consequence of the regularization term $\frac{\lambda}{2} \sum_{i=1}^N  \bm{h}_{i}^\top \bm{h}_i$ in the training objective (see \eqref{eq:obj_train_weighted}), which implicitly puts a Gaussian prior on the hidden units. When the training of feature map is done correctly, one expects the latent variables to be normally distributed around zero~\cite{suykens_least_2002}. Gaussian distributed latent variables are essential for having a \emph{continuous} and \emph{smooth} latent space, allowing easy interpolation.  This property is also essential for VAEs and  was studied in \cite{kingma_auto-encoding_2013}, where a regularization term, in the form of the Kullback-Leibler divergence between the encoder's distribution and a unit Gaussian as a prior on the latent variables was used. When training a non-robust generative model in the presence of outliers, the contamination can severely distort the distribution of the latent variables. This effect is seen in Figure~\ref{fig:histogramOutliers}, where a discontinuous and skewed distribution is visible. %
\subsection{Algorithm}
\label{sec:algorithm}
We propose to use the above described reweighting step within the Gen-RKM framework~\cite{GENRKM}. The algorithm is flexible to incorporate both kernel-based, (deep) neural network and Convolutional based models within the same setting, and is capable of jointly learning the feature maps and latent representations. The Gen-RKM algorithm consists out of two phases: a training phase and a generation phase which occurs one after another.
In the case of explicit feature maps, the training phase consists of determining the parameters of the explicit feature and pre-image map together with the hidden units $\{\bm{h}_i\}_{i=1}^N$. 

We propose an adapted algorithm of \cite{GENRKM} with an extra re-weighting step wherein the system in \eqref{eq:weightedEIG} is solved. Furthermore, the reconstruction error is weighted to reduce the effect of potential outliers on the pre-image maps.  The loss function now becomes:
\begin{equation}
    \label{eq:combined_trainigEqWe}
    \begin{aligned}
    \underset{\bm{\theta},\bm{\zeta}}{\min} \enskip  \mathcal{J}_{c}^{\bm{D}}(\bm{\theta},\bm{\zeta})  = &\enskip  \mathcal{J}_t^{\bm{D}} + \frac{c_{\mathrm{stab}}}{2} (\mathcal{J}_{t}^{\bm{D}})^2 + \frac{c_{\mathrm{acc}}}{N}\sum_{i=1}^N  \bm{D}_{ii} \mathcal{L}(\bm{x}_i,\psi_{\bm{\zeta}} (\varphi_{\bm{\theta}} (\bm{x}_i))),
    \end{aligned}
\end{equation}
where $c_{stab}\in \mathbb{R}^{+}$ is a stability constant \cite{suykens_deep_2017} and $c_{\mathrm{acc}}\in \mathbb{R}^+$ is a regularization constant to control the stability with reconstruction accuracy. In the experiments, the loss function is equal to the mean squared error (MSE), however other loss functions are possible. The generation algorithm is the same as in \cite{GENRKM}.%

\begin{table}[ht]
	\caption{FID Scores~\cite{Heusel2017} over 10 iterations for 4000 randomly generated samples when the training data is contaminated with 20\% outliers. (smaller is better).}
	\label{Table:fid}
	\centering
	\begin{tabular}{|c|c|c|c|c|}
		\hline
		\multirow{2}{*}{\textbf{Dataset}} & \multicolumn{4}{c|}{\textbf{FID score}}                                       \\ \cline{2-5}
	& \textbf{VAE}& \textbf{$\beta$-VAE} ($\beta=3$) & \textbf{RKM} & \textbf{Rob Gen-RKM} \\ \hline
		\multirow{1}{*}{\textbf{MNIST}}
		 & 142.54$\pm$0.73 & 187.21$\pm$0.11 & 134.95$\pm$1.61 & \textbf{87.32$\pm$1.92}\\ \hline
	\multirow{1}{*}{\textbf{F-MNIST}}
	 & 245.84$\pm$0.43	 & 291.11$\pm$1.6 & 163.51$\pm$1.24 & \textbf{153.32$\pm$0.05}\\ \hline
	 \multirow{1}{*}{\textbf{SVHN}}
	 & 168.21$\pm$0.23 & 234.87$\pm$1.45 & 112.45$\pm$1.4 & \textbf{98.14 $\pm$1.2}\\ \hline
	 	 \multirow{1}{*}{\textbf{CIFAR-10}}
	 & 201.21$\pm$0.71 & 241.23$\pm$0.34 & 187.08$\pm$0.58 & \textbf{132.6$\pm$0.21} \\ \hline
	\multirow{1}{*}{\textbf{Dsprites}}
	& 234.51$\pm$1.10	& 298.21$\pm$1.5 & 182.65$\pm$0.57 & \textbf{160.56$\pm$0.96}  \\ \hline
	\multirow{1}{*}{\textbf{3Dshapes}}
		& 233.18$\pm$0.94 & 252.41$\pm$0.38 & 177.29$\pm$1.60 & \textbf{131.18$\pm$1.45}  \\ \hline
		  \end{tabular}
\end{table}

\section{Experiments}
\label{sec:Experiments}
In this section, we evaluate the robustness of the weighted Gen-RKM on the MNIST, Fashion-MNIST (F-MNIST), CIFAR-10, SVHN, Dsprites and 3Dshapes dataset\footnote{\scriptsize{\url{http://yann.lecun.com/exdb/mnist/}, \url{https://github.com/zalandoresearch/fashion-mnist}, \url{https://github.com/deepmind/dsprites-dataset}, \url{https://github.com/deepmind/3d-shapes},  \url{https://www.cs.toronto.edu/~kriz/cifar.html}, \url{http://ufldl.stanford.edu/housenumbers/}}}. The last two datasets will be used in disentanglement experiments since they include the ground truth generating factors which are necessary to quantify the performance. Training of the robust Gen-RKM is done using the algorithm proposed in Section \ref{sec:algorithm}, where we take $N_\mathrm{{MCD}} = \lfloor N \times 0.75 \rfloor$ and $\alpha = 0.975$ for the Chi-squared distribution (see \eqref{eq:weights}). Afterwards we compare with the standard Gen-RKM~\cite{GENRKM}, VAE and $\beta$-VAE. The models have the same encoder/decoder architecture, optimization parameters and are trained until convergence. Information on the training settings and model architectures is given in the Appendix.\\
\begin{figure*}[]
    \centering
    \begin{subfigure}{0.45\textwidth}
		\centering
		\includegraphics[width=1\textwidth]{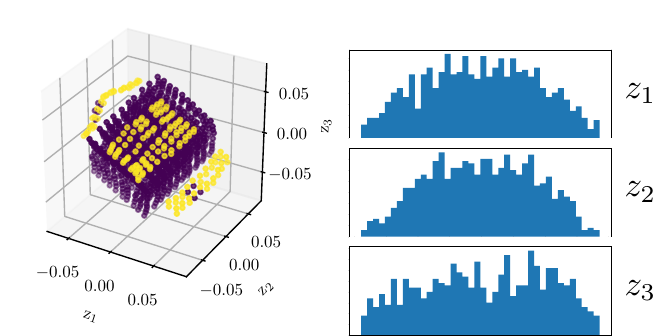}
		\caption{Gen-RKM}
	\end{subfigure}
	\begin{subfigure}{0.45\textwidth}
		\centering
		 \includegraphics[width=1\textwidth]{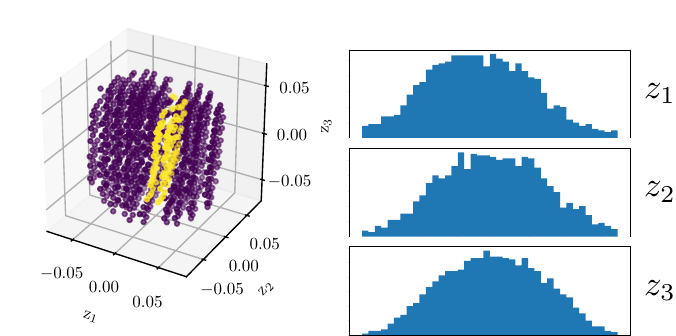} 
		\caption{Robust Gen-RKM}
	\end{subfigure}
    \caption[]{Illustration of disentanglement on the 3DShapes dataset. Clean data is depicted in purple, outliers in yellow. The training subset is contaminated with a third generating factor (20\% of the data is considered as outliers). The outliers are down-weighted in the robust Gen-RKM, which moves them to the center.}
    \label{fig:disang}
\end{figure*}
\textbf{Generation and Denoising}: Figure \ref{fig:RandomGeneration} shows the generation of random images when models were trained on the contaminated MNIST dataset. The contamination consists of adding Gaussian noise $\mathcal{N}(0.5,0.5)$ to 20\% of the data. The images are generated by random sampling from a fitted Gaussian distribution on the learned latent variables. As we can see, when using a robust training procedure, the model does not encode the noisy images. As a consequence, no noisy images are generated and the generation quality is significantly better. This is also confirmed by the Fr\'echet Inception Distance (FID) scores~\cite{Heusel2017} in Table~\ref{Table:fid}, which quantifies the quality of generation. The robust Gen-RKM clearly outperforms the other methods when the data has contamination. Moreover VAEs are known to generate samples closer to the mean of dataset. This negatively affects the FID scores which also takes into account the diversity within the generated images. The scores for $\beta$-VAE are worst due to the inherent emphasis on imposing a Gaussian distribution on latent variables trading-off with the reconstruction quality~\cite{higgins2017beta}. The classical RKM performs slightly better than VAE and its variant. This is attributed to the presence of kernel PCA during training, which is often used in denoising applications and helps to some extent in dealing with contamination in the dataset.

Next, we use generative models in the denoising experiment. Image denoising is accomplished by projecting the noisy test set observations on the latent space, afterwards projecting back to the input space. Because there is a latent bottleneck, the most important features of the images are retained while insignificant features like noise are removed. Figure~\ref{fig:Densoing} shows an illustration of robust denoising on the MNIST dataset. The robust Gen-RKM does not encode the noisy images within the training procedure. Consequently, the model is capable of denoising the out-of-sample test images. When comparing the denoising quality on the full test set (5000 images sampled uniformly at random), the Mean Absolute Error (MAE) of Gen-RKM: $\mathrm{MAE} = 0.415$ and VAE: $\mathrm{MAE} = 0.434$ is much higher than the robust version: $\mathrm{MAE} = 0.206$. The experiments show that basic generative models like Gen-RKM and VAE are highly affected by outliers, while the robust counterpart can cope with a significant fraction of contamination.
\begin{figure*}[h]
    \centering
  \includegraphics[width=0.8\textwidth]{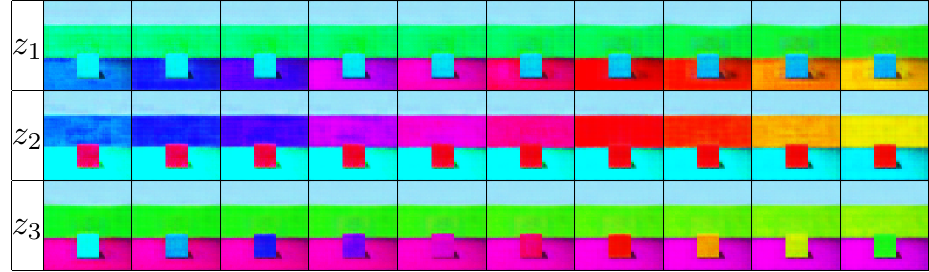}
    \caption[]{Illustration of latent traversals along the 3 latent dimensions for 3DShapes dataset using the robust Gen-RKM model. The first, second and third row distinctly captures the floor-hue, wall-hue and object-hue respectively while keeping other generative factors constant. }%
    \label{fig:3Dshapes}
\end{figure*}
\begin{table}[h]
	\caption{Disentanglement Metric on DSprites and 3D Shapes dataset. The training subset is contaminated with extra generating factors (20\% of the data is considered as outliers). The framework of \cite{eastwood2018a} with Lasso and Random Forest regressor~\cite{eastwood2018a} is used to evaluate the learned representation. For disentanglement and completeness higher score is better, for informativeness, lower is better.}
	\label{Table:disang}
	\begin{adjustbox}{width=\columnwidth,center}
	\begin{tabular}{|c|c|c|c|c|c|c|c|c|}
		\hline
	\multirow{2}{*}{\textbf{Dataset}}	&      \multirow{2}{*}{$h_{dim}$}                 &    \multirow{2}{*}{\textbf{Algorithm}}                                       & \multicolumn{3}{c|}{\textbf{Lasso}} & \multicolumn{3}{c|}{\textbf{Random Forest}}                                                                             \\
		\cline{4-9}
		 &              &                         & \textbf{Disent.}                    & \textbf{Comple.}                            & \textbf{Inform.} & \textbf{Disent.} & \textbf{Comple.} & \textbf{Inform.} \\  \hline
	\multirow{2}{*}{{\textbf{DSprites}}}	& \multirow{2}{*}{2}&          $\beta$-VAE ($\beta=3$)                                       & 0.19                                & 0.16                                        & 6.42    & 0.13   & 0.32    &  \textbf{1.39}  \\  &  & Gen-RKM           & 0.07                               & 0.07                                       & \textbf{5.82}             & 0.25             & 0.27             & {5.91}            \\
		&                       & Rob Gen-RKM                                       & \textbf{0.21}                                & \textbf{0.21}                                        & 9.13    & \textbf{0.36}   & \textbf{0.38}    & 5.95   \\ \hline
		\multirow{2}{*}{{\textbf{3DShapes}}}
		& \multirow{2}{*}{3}    & $\beta$-VAE ($\beta=3$)                                       & 0.24                               & 0.28                                        &   \textbf{2.72}  & 0.12   & 0.13    &  2.15  \\ &        & Gen-RKM                                   & 0.14                                & 0.14                                   & {3.03}             & 0.15    & 0.15    & 1.09   \\
		&                       & Rob Gen-RKM                                       & \textbf{0.47}                              & \textbf{0.49}                                      & 3.13    & \textbf{0.44}            & \textbf{0.45}             & \textbf{1.02}     \\  \hline
	\end{tabular}
\end{adjustbox}
\end{table}
\textbf{Effect on Disentanglement:}
In this experiment, contamination is an extra generating factor which is not present in the majority of the data. The goal is to train a disentangled representation, where the robust model only focuses on the most prominent generating factors. We subsample a `clean' training subset which consists of cubes with different floor, wall and object hue. The scale and orientation are kept constant with minimum scale and $0^{\circ}$ orientation respectively. Afterwards, the training data is contaminated by cylinders with maximum scale at $30^{\circ}$ orientation (20\% of the data is considered as outliers). The training data now consist out of 3 `true' generating factors (floor, wall and object hue) which appear in the majority of the data and 3 `noisy' generating factors (object, scale and orientation) which only occur in a small fraction. To illustrate the effect of the weighting scheme, Figure~\ref{fig:disang} visualizes the latent space of the (robust) Gen-RKM model. The classical Gen-RKM encodes the outliers in the representation, which results in a distorted Gaussian distribution of the latent variables. This is not the case for the robust Gen-RKM, where the outliers are downweighted. An illustration of latent traversals along the 3 latent dimensions using the robust Gen-RKM model is given in Figure~\ref{fig:3Dshapes}, where the robust model is capable of disentangling the 3 `clean' generating factors.

To quantify the performance of disentanglement, we use the proposed framework\footnote{Code and dataset available at \url{https://github.com/cianeastwood/qedr}} of \cite{eastwood2018a}, which consists of 3 metrics: disentanglement, completeness and informativeness. The framework could be used when the \emph{ground-truth latent structure is available}, which is the case for 3Dshapes and DSprites dataset. The results are shown in Table~\ref{Table:disang}, where the robust method outperforms the Gen-RKM. The above experiment is repeated on the DSprites dataset. The `clean' training subset consists of ellipse shaped datapoints with minimal scale and $0^{\circ}$ angle at different $x$ and $y$ positions. Afterwards, the training data is contaminated with a random sample of different objects at larger scales, different angles at different $x$ and $y$ positions. The training data now consist out of 2 `true' generating factors ($x$ and $y$ positions) which appear in the majority of the data and 3 `noisy' generating factor (orientation, scale and shape) which only occur in a small fraction. In addition to RKM, the results of $\beta$-VAE are shown in Table~\ref{Table:disang}.
\section{Conclusion}
\label{sec:Conclusion}
Using a weighted conjugate feature duality, a RKM formulation for weighted kernel PCA is proposed. This formulation is used within the Gen-RKM training procedure to fine-tune the latent space using a weighting function based on the MCD. Experiments show that the weighted RKM is capable of generating denoised images in spite of contamination in the training data. Furthermore, being a latent variable model, robust Gen-RKM preserves the disentangled representation. Future work consists of exploring various robust estimators and other weighting schemes to control the effect of sampling bias in the data. 

\subsubsection*{Acknowledgments}
\footnotesize{
EU: The research leading to these results has received funding from
the European Research Council under the European Union's Horizon
    2020 research and innovation program / ERC Advanced Grant E-DUALITY
    (787960). This paper reflects only the authors' views and the Union
    is not liable for any use that may be made of the contained information.
Research Council KUL: Optimization frameworks for deep kernel machines C14/18/068
Flemish Government:
FWO: projects: GOA4917N (Deep Restricted Kernel Machines:
        Methods and Foundations), PhD/Postdoc grant
Impulsfonds AI: VR 2019 2203 DOC.0318/1QUATER Kenniscentrum Data
        en Maatschappij
Ford KU Leuven Research Alliance Project KUL0076 (Stability analysis
    and performance improvement of deep reinforcement learning algorithms).}%

\section*{Appendix}
Table \ref{tab:arch} shows the  details  on  training settings used in this paper. The PyTorch library in Python was used  with a  8GB NVIDIA QUADRO P4000 GPU.
\begin{table}[h]
\caption{Model architectures. All convolutions and transposed-convolutions are with stride 2 and padding 1. Unless stated otherwise, layers have Parametric-RELU ($\alpha = 0.2$) activation functions, except output layers of the pre-image maps which have sigmoid activation functions. $N_{\mathrm{sub}} \leq N$  is the training subset size, $s$ the latent space dimension and $m$ the minibatch size.  \label{tab:arch}}
\centering
	\begin{tabular}{l c l c c c}
		\specialrule{.1em}{.05em}{.05em}
		\textbf{Dataset}          & \multicolumn{1}{l}{\textbf{Optimizer}}         &                    \multicolumn{2}{c}{\textbf{Architecture}}   &  \multicolumn{2}{c}{\textbf{Parameters}}                                   \\
		& \multicolumn{1}{c}{(Adam)} &                    & & &    \\
		\specialrule{.1em}{.05em}{.05em}
		\multirow{3}{*}{\pbox{3cm}{MNIST/F-MNIST/\\CIFAR-10/SVHN}}    & \multirow{3}{*}{1e-4}      & Feature-map (fm)   & \pbox{20cm}{Conv 32$\times$4$\times$4;         \\  Conv 64$\times$4$\times$4; \\ FC 228 (Linear)}    & \pbox{10cm}{$N_{\mathrm{sub}}$}  &    \pbox{10cm}{3000}             \\
		&                            & Pre-image map      & reverse of fm    & $s$ & 10                                      \\
		&                            & Latent space dim.      & 10      & $m$ & 200                                         \\
		\hline
		\multirow{3}{*}{\pbox{3cm}{Dsprites/3DShapes}} & \multirow{3}{*}{1e-4}       & Feature-map (fm)   & \pbox{20cm}{Conv 20$\times$4$\times$4;                                                        \\ Conv 40$\times$4$\times$4;\\ Conv 80$\times$4$\times$4;\\ FC 228 (Linear)}     & $N_{\mathrm{sub}}$  &    1024/1200     \\
		&                            & Pre-image map      & reverse of fm   & $s$ & 2                                                       \\
		&                            & Latent space dim.  & $2/3$  & $m$ & 200                                                   \\
		\hline
	\end{tabular}
\end{table}

\FloatBarrier

\bibliographystyle{ieeetr}
\bibliography{IEEEabrv,References}

\end{document}